\documentclass[11pt,a4paper]{article}

\usepackage[margin=1in]{geometry}
\usepackage[T1]{fontenc}
\usepackage[utf8]{inputenc}
\usepackage{lmodern}
\usepackage{microtype}
\usepackage{setspace}
\usepackage{graphicx}
\usepackage{booktabs}
\usepackage{array}
\usepackage{tabularx}
\usepackage{caption}
\usepackage[round,authoryear]{natbib}
\usepackage[hidelinks]{hyperref}
\usepackage{enumitem}
\usepackage{amsmath,amssymb}
\usepackage{xcolor}
\DeclareMathOperator{\Logit}{logit}

\graphicspath{{figures/}}
\hypersetup{colorlinks=true,linkcolor=blue!55!black,citecolor=blue!55!black,urlcolor=blue!55!black}

\newcommand{\suite}[1]{\texttt{#1}}
\newcommand{\family}[1]{\texttt{#1}}

\title{\textbf{Arithmetic OOD Failure Unfolds in Stages in Minimal GPTs}}
\author{Seine A.~Shintani$^{1,2,3}$}
\date{March 2026}

\begin{document}
\maketitle

\begin{center}
\small
$^{1}$ Department of Biomedical Sciences, College of Life and Health Sciences, Chubu University, Kasugai, Aichi 487-8501, Japan\\
$^{2}$ Center for Mathematical Science and Artificial Intelligence, Chubu University, Kasugai, Aichi 487-8501, Japan\\
$^{3}$ Institute for Advanced Research, Nagoya University, Nagoya, Aichi 464-8601, Japan\\[0.4em]
ORCID: 0000-0002-1084-2549\\
Correspondence: \href{mailto:s-shintani@fsc.chubu.ac.jp}{s-shintani@fsc.chubu.ac.jp}
\end{center}

\begin{abstract}
Arithmetic benchmarks are often reduced to a single held-out score, but that score can conflate qualitatively different failures. We study a controlled minimal GPT trained on exhaustive 2-digit addition, where all local digit transitions are already present in training, and ask why 3-digit generalization still fails. The failure is staged. First, there is a layout barrier: a learned absolute-position model collapses under a pure 3-digit layout shift, and mixed-layout exposure is the only intervention that materially weakens this barrier. Second, after layout repair, the hundreds position behaves like a carry flag rather than a semantic hundreds digit; targeted carry probes reverse the relevant logit margin, whereas a matched extra-data control does not. Third, after carry repair, the main remaining bottleneck is conditional recomposition: high-conditioned tail data outperforms a matched control, high-only data, and tail-only data on all true-3-digit suites, and the same ordering reappears in a larger 2-layer bridge experiment. The residual errors after recomposition are then overwhelmingly tens-only, and a separate 10-seed late-stage study shows that a sign-aware tens repair raises exact match on the hardest thousands-carry suite from 0.664 to 0.822. We therefore provide an experimentally testable decomposition of arithmetic OOD failure into layout, carry-semantics, recomposition, and late tens-residual stages.
\end{abstract}


\section{Introduction}

Addition is a useful stress test for sequence models because it is exact, synthetic, and exhaustively generable. These properties make it unusually well suited to the study of generalization. The training distribution can be written down completely, test-time shifts can be introduced one at a time, and resulting failures can be read at the level of individual digits and carries. More broadly, algorithmic tasks have long been used to study systematic generalization in neural sequence models, from early program-execution and arithmetic settings \citep{zaremba2014execute,kaiser2015neuralgpu} to later Transformer-based models designed to improve symbolic extrapolation \citep{dehghani2019universal}. Recent work has shown that small Transformers can exhibit delayed generalization on algorithmic data \citep{power2022grokking,nanda2023grokkingmi}, that arithmetic performance depends strongly on formatting and representation choices \citep{lee2024teaching,zhou2024lengthnotrobust,cho2024position,mcleish2024abacus,sabbaghi2024symmetry,cho2025operandcount}, and that length generalization itself deserves both formal and empirical analysis \citep{huang2025formal,abbe2024scratchpad}. Diagnostics on larger language models, meanwhile, show that apparently strong arithmetic behavior can remain fragile under notation changes, operand reordering, and length shifts \citep{yan2025additiondiagnostic}.

What is still often missing is a careful account of \emph{which} failure is being measured. A single exact-match score can combine several qualitatively different problems. A model may already know every local digit rule needed for later addition, yet still fail when the same low-range numbers are written in a different layout. Once that representational problem is weakened, the model may still use an upper position as a carry indicator instead of as a place-value digit. Even after the upper digits become correct, the model may still attach the wrong lower tail to them. If those barriers are folded into one number, ``arithmetic generalization'' becomes too coarse to explain what changed and why.

This paper addresses a sharper question. When a minimal GPT trained on exhaustive 2-digit addition fails on 3-digit cases, what fails first, what fails next, and what remains after each targeted repair? We show that the answer is staged. The first barrier is a layout barrier. The second is a semantic barrier: the hundreds position initially behaves like a carry flag. The third is a compositional barrier: after carry repair, the model often knows the correct upper state but still fails to bind the correct lower tail. The final major residual is much narrower and appears as a carry-conditioned tens-sign error.

The value of this sandbox is that it lets us identify which barrier is active at each stage of failure. Using a tensorized mirror of microGPT \citep{karpathy2026microgpt}, we combine exhaustive data generation, tightly controlled shifts, matched data packs, and repeated seed sweeps. This turns arithmetic OOD failure from a single held-out score into a sequence of concrete and testable claims.

Across the paper, that sequence becomes stable and interpretable. A five-seed position-family comparison shows that structured positional bias alone does not remove the earliest exact-match barrier; mixed-layout exposure does. A matched five-seed carry-probe study then shows that the hundreds position behaves like a carry flag, and that the effect is content-specific rather than a by-product of extra optimization. A matched binding ablation shows that once carry semantics improve, the next bottleneck is conditional recomposition. The same ordering reappears in a larger 2-layer, width-32 bridge experiment, which connects the analytic 1-layer study to the later residual analysis. Finally, a 10-seed late-stage study shows that once recomposition is largely repaired, the remaining error is a narrow carry-conditioned tens residual. Taken together, these results suggest that arithmetic failure in this sandbox is not monolithic. It appears in stages, and those stages can be isolated experimentally.

\begin{figure}[t]
\centering
\includegraphics[width=\linewidth]{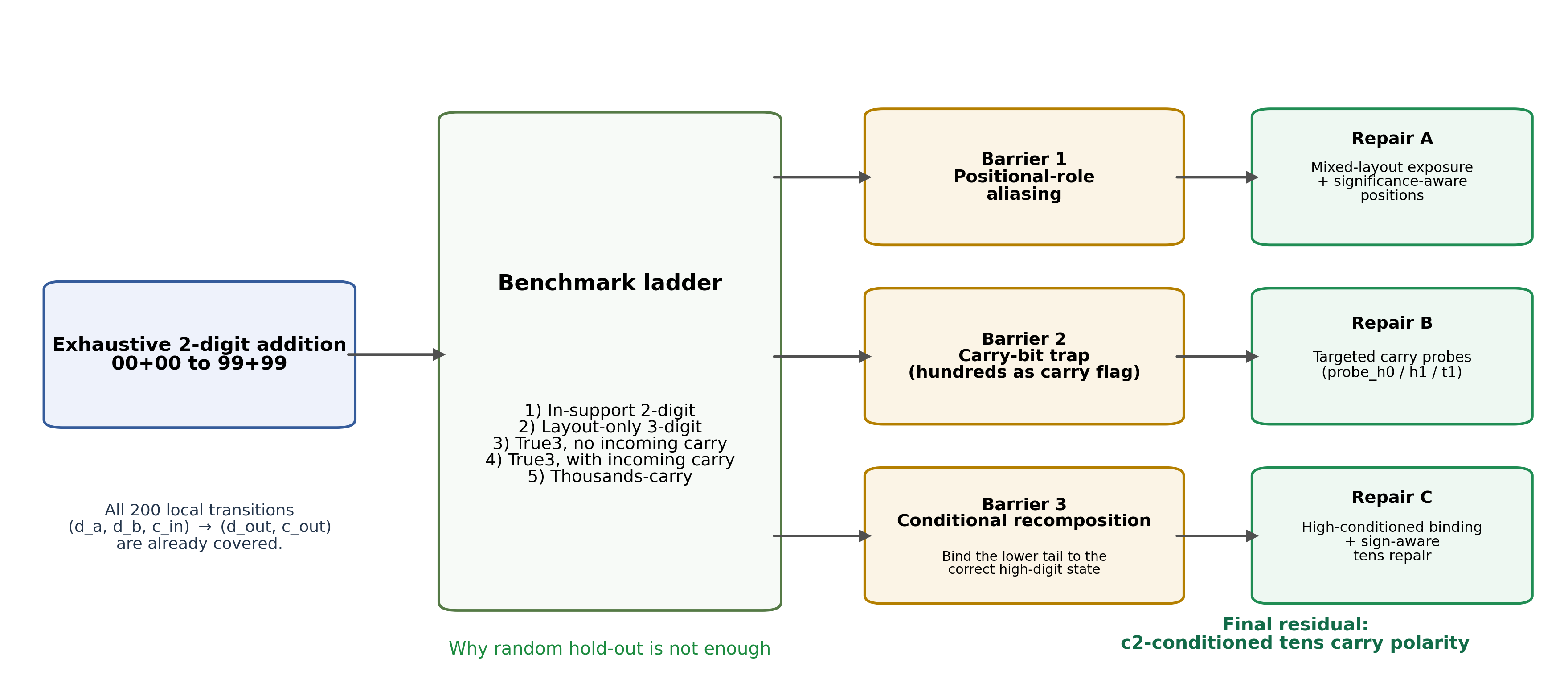}
\caption{\textbf{Overview of the benchmark ladder and repair program.} Exhaustive 2-digit addition already covers all 200 local digit transitions, so later 3-digit failures cannot be explained by missing local rules alone. The evaluation suites are ordered so that each new suite adds one qualitatively different source of difficulty. The repair families are matched to the active barrier rather than being generic augmentation.}
\label{fig:overview}
\end{figure}

\section{Related Work}

Algorithmic tasks have long served as controlled settings for studying systematic generalization. Earlier recurrent models were evaluated on symbolic execution and arithmetic \citep{zaremba2014execute,kaiser2015neuralgpu}, and later architectures such as the Universal Transformer were motivated in part by the difficulty that feed-forward sequence models have with extrapolation on simple symbolic problems \citep{dehghani2019universal}. Our work shares that preference for tightly controlled synthetic data, but it asks a different question. Rather than asking only whether a held-out score improves, we ask which barrier is being crossed when it improves.

A second line of work studies arithmetic and length generalization directly. Formatting and curriculum can change arithmetic performance dramatically \citep{lee2024teaching}. Position design matters as well: in the broader Transformer literature, relative positions and linear attention biases have been used to improve the handling of longer contexts \citep{shaw2018relative,press2022alibi}. In arithmetic-specific settings, recent work has used task-structured position coupling \citep{cho2024position}, digit-aware embeddings and related architectural changes \citep{mcleish2024abacus}, explicit structural symmetry \citep{sabbaghi2024symmetry}, and multi-level task-specific scratchpads \citep{cho2025operandcount} to improve extrapolation. More broadly, arithmetic has become a model system for length-generalization research \citep{zhou2024lengthnotrobust,huang2025formal,abbe2024scratchpad}. These studies motivate our comparisons, but our objective is different. We do not try to reproduce their full systems or claim a new state of the art. Instead, we use a common sandbox to ask which failure remains under a tightly controlled data regime.

A third thread concerns mechanistic analysis of arithmetic in Transformers. \citet{quirke2024addition} analyze addition circuits in a one-layer Transformer, and \citet{quirke2024arithmetic} extend that line of work. Our study is complementary. We use behavior, logits, and attention summaries to isolate stage-specific errors across matched interventions and seed sweeps, but we stop short of a full circuit reconstruction.

Finally, work on larger language models shows that arithmetic competence can remain brittle even when headline scores look strong. \citet{yan2025additiondiagnostic} show that changes in notation, digit scale, and algebraic structure can expose shallow pattern matching. Our paper is narrower in scope, but it offers a vocabulary for describing what a brittle arithmetic score may actually contain: a mixture of layout confusion, semantic misassignment, failed recomposition, and a narrower late residual that only becomes visible once earlier barriers have been weakened.

\section{Method}

\subsection{Controlled minimal GPT sandbox}

The experimental platform is a minimal decoder-only Transformer inspired by the microGPT educational implementation \citep{karpathy2026microgpt}. Like standard Transformer models, it relies on self-attention and therefore needs positional information or positional bias to distinguish order \citep{vaswani2017attention,shaw2018relative,press2022alibi}. Arithmetic is a useful place to study positional design because digit meaning depends directly on place value. The tensorized mirror used here preserves the same broad modeling assumptions as the original microGPT code while enabling batched GPU training and repeated seed sweeps.

The purpose of the sandbox is control rather than scale realism. In this setting, the data can be generated exhaustively, distribution shifts can be introduced one at a time, interventions can be volume matched, and residual errors can be tracked from one stage to the next. That level of control lets us replace one monolithic ``3-digit accuracy'' number with a barrier-by-barrier analysis.

The common base is exhaustive 2-digit addition: all $100 \times 100 = 10{,}000$ ordered input pairs formatted as zero-padded character strings. Mixed-layout families append the low-range 3-digit rendering of the same 10{,}000 ordered pairs. Tokenization is character-level over digits and symbols, with context length 13 so that both 2-digit and 3-digit formats fit into a single autoregressive window including the initial \texttt{<BOS>} token. Unless otherwise noted, we use batch size 256, learning rate $3 \times 10^{-3}$, zero weight decay, and 1{,}000 evaluation examples per suite per seed.

The position-family, carry-probe, and analytic binding studies use a 1-layer, width-16, 4-head model. The bridge experiment and the late-stage tens-residual study use a 2-layer, width-32, 4-head model.

\subsection{Tasks and evaluation suites}

All tasks are character-level next-token prediction over zero-padded addition strings. A 2-digit training example has the form \texttt{49+07=056}, while a true 3-digit evaluation example has the form \texttt{201+760=0961}. Padding is deliberate. It makes place value explicit, keeps layout controlled across suites, and separates arithmetic meaning from uncontrolled length variation.

We write $c_2$ for the carry from the tens column into the hundreds column and $c_3$ for the carry from the hundreds column into the thousands column. These carries matter because the late residual is not merely ``a tens error'' in the abstract; it depends systematically on the carry state entering the upper digits.

Exact match means that the full generated answer string is correct. Teacher-forced token accuracy is a softer metric: given the true prefix, it measures how often the model predicts the correct next character. For later-stage analysis we also use \texttt{high2\_correct}, \texttt{low2\_correct}, and $P(\mathrm{exact}\mid\mathrm{high2\ correct})$ to distinguish ``the upper state is right'' from ``the whole answer is right.''

\begin{table}[t]
\centering
\small
\caption{\textbf{Evaluation suites.} The suites are ordered so that the first failure can be interpreted cleanly. The layout-only suite changes formatting but not value range. The later suites keep 3-digit layout and add semantic difficulty through carry structure.}
\label{tab:suites}
\begin{tabularx}{\linewidth}{>{\raggedright\arraybackslash}p{3.1cm} >{\raggedright\arraybackslash}X >{\raggedright\arraybackslash}p{3.0cm}}
\toprule
Human-readable label & What it tests & Example target string \\
\midrule
2-digit in-support & Standard 2-digit format seen during training (code name: \suite{in\_support\_2digit}) & \texttt{49+07=056} \\
3-digit layout-only, low range & Same 0--99 value range, but written in 3-digit layout (code name: \suite{layout\_shift\_only\_lowrange3}) & \texttt{049+007=0056} \\
True 3-digit, no incoming carry & Genuine 3-digit addition with no incoming carry to hundreds ($c_2=0$; code name: \suite{true3\_hundreds\_no\_incarry}) & \texttt{241+320=0561} \\
True 3-digit, incoming carry & Genuine 3-digit addition with an incoming carry to hundreds ($c_2=1$; code name: \suite{true3\_hundreds\_with\_incarry}) & \texttt{289+173=0462} \\
Thousands-carry & Carry into a new thousands digit ($c_3=1$; includes both $c_2=0$ and $c_2=1$ cases; code name: \suite{true3\_thousands\_carry}) & \texttt{782+289=1071} \\
\bottomrule
\end{tabularx}
\end{table}

\subsection{Why random hold-out is not enough}

The benchmark is built around one central observation: exhaustive 2-digit addition already saturates all local digit transitions. Every local rule of the form
\[
(d_a,d_b,c_{in}) \rightarrow (d_{out},c_{out})
\]
appears somewhere in the exhaustive 2-digit dataset. When the model later fails on 3-digit arithmetic, the failure therefore cannot be blamed on missing local rules alone.

This changes how held-out accuracy should be interpreted. A random 80/20 split inside the 2-digit format can look like an OOD test, but it is largely solved by complete local coverage. In our split analysis, a local-rule reconstruction baseline achieves perfect coverage and perfect accuracy on the IID random hold-out split, yet fails on structured splits such as \suite{units\_pair\_ood}, \suite{tens\_carry\_ood}, and \suite{carry\_chain\_ood}. The methodological lesson is simple: once local coverage is saturated, informative arithmetic evaluation requires structured barriers rather than random masking.

\begin{table}[t]
\centering
\small
\caption{\textbf{Why IID random hold-out is insufficient.} ``Local'' refers to a baseline that reconstructs answers from observed local digit transitions; ``Sym.'' adds commutative symmetry. The IID split is already solved by complete local coverage, whereas structured splits expose qualitatively different barriers.}
\label{tab:splits}
\begin{tabularx}{\linewidth}{>{\raggedright\arraybackslash}p{3.1cm} c c c >{\raggedright\arraybackslash}X}
\toprule
Split & Local coverage & Local accuracy & Sym. accuracy & Main missing structure \\
\midrule
IID 80/20 random hold-out & 1.000 & 1.000 & 1.000 & none \\
Units-pair OOD & 0.000 & 0.000 & 0.850 & unseen ones-pair combinations \\
Tens-carry OOD & 0.000 & 0.000 & 0.701 & unseen tens+carry combinations \\
Carry-chain OOD & 0.000 & 0.000 & 0.000 & sustained carry propagation \\
Ordered commutativity OOD & 0.091 & 0.091 & 1.000 & compositional symmetry \\
\bottomrule
\end{tabularx}
\end{table}

\subsection{Comparison families}

The empirical program has five linked parts.

First, we compare positional families at the earliest barrier. The families include a learned absolute-position baseline, a mixed-layout absolute-position family that sees the 3-digit rendering of the same low-range sums during training, a coupled-significance family, a digit-aware family, a symmetry-aware family, and a no-position family. The coupled-significance, digit-aware, and symmetry-aware variants are simplified inspired-by realizations of prior ideas \citep{cho2024position,mcleish2024abacus,sabbaghi2024symmetry}; the goal is controlled inductive-bias comparison rather than exact reimplementation.

Second, we run a matched carry-probe study on a mixed-layout coupled-significance base. Here we compare the base family, a matched extra-data control (\family{ctrl\_dup}), and a targeted probe family (\family{probe\_all}). The matched control receives the same additional training volume as the targeted family but with in-support duplicate content. This lets us test whether weakening the carry-flag trap depends on what is added or merely on how much is added.

Third, from a shared carry-repaired base in the 1-layer, width-16 model, we run a matched binding ablation. The final matched pack is varied across four families: a matched control (\family{ctrl3}), a high-only pack, a tail-only pack, and a high-conditioned tail pack (\family{tailhigh}). This comparison asks what ingredient is actually needed once the carry-flag trap has already been weakened.

Fourth, we repeat that same conditional-binding comparison in a 2-layer, width-32 model. This bridge experiment is not a different conceptual stage. It tests whether the same ordering reappears at the capacity used later for the final residual study.

Fifth, we run a separate 10-seed late-stage confirmation study in the larger 2-layer, width-32 model. It compares \family{ctrl4}, \family{tensboundary}, and \family{tenspolarity} to test whether the last major residual is a carry-conditioned tens-sign error.

\subsection{Metrics and robustness}

We report exact match, format validity, teacher-forced token accuracy, position-wise digit accuracy, \texttt{high2\_correct}, \texttt{low2\_correct}, and $P(\mathrm{exact}\mid\mathrm{high2\ correct})$ as a recomposition metric. The position-family, carry-probe, analytic binding, and large-capacity bridge studies each use 5 seeds. The final tens-residual confirmation study uses 10 seeds and includes paired-seed comparisons on the hardest suite.

\section{Results}

\subsection{The first barrier is a layout barrier}

The first result answers a simple question: can better positional inductive bias alone remove the earliest exact-match failure? The five-seed comparison in Table~\ref{tab:posfam} shows that it cannot. The absolute-position baseline learns the in-support 2-digit task well (0.920 exact), yet collapses when the same 0--99 sums are merely written in 3-digit layout (0.0008 exact on \suite{layout\_shift\_only\_lowrange3}). Mixed-layout exposure is the only intervention that materially weakens this first barrier, lifting exact match on the layout-only suite to 0.4426.

The more structured positional families tell a subtler story. The coupled-significance and symmetry-aware variants improve teacher-forced token accuracy on the layout-only suite from 0.201 to 0.332 and 0.357, respectively, but exact match remains essentially zero. These results do not imply that positional bias is irrelevant. The coupled-significance and symmetry-aware families clearly improve token-level formatting and role stability. What Table~\ref{tab:posfam} shows is narrower: in this sandbox, positional bias alone does not substitute for seeing the shifted layout itself.

\begin{table}[t]
\centering
\small
\caption{\textbf{Five-seed position-family comparison.} Only the mixed-layout family sees the 3-digit rendering of the same low-range sums during training; the other families test whether positional inductive bias alone weakens the first barrier. Exact-match means use constrained decoding. The final column reports exact match on the no-incoming-carry true-3-digit suite.}
\label{tab:posfam}
\begin{tabular}{>{\raggedright\arraybackslash}p{3.0cm}cccc}
\toprule
Family & 2-digit & Layout-only & Layout token acc. & True 3-digit (no incarry) \\
\midrule
Absolute & 0.9200 & 0.0008 & 0.201 & 0.0002 \\
Mixed-layout absolute & 0.5886 & \textbf{0.4426} & \textbf{0.854} & 0.0000 \\
Coupled significance & 0.7990 & 0.0000 & 0.332 & \textbf{0.0022} \\
Digit-aware & 0.4288 & 0.0002 & 0.133 & 0.0000 \\
Symmetry-aware & 0.7482 & 0.0000 & 0.357 & 0.0006 \\
No position & 0.0944 & 0.0006 & 0.081 & 0.0002 \\
\bottomrule
\end{tabular}
\end{table}

\subsection{After layout repair, the hundreds position behaves like a carry flag}

Once layout is stabilized, the next question is semantic: does the model treat the hundreds position as a place-value digit, or as a marker that a carry arrived from below? The matched carry-probe study favors the second description. In the base family, exact match on the three true-3-digit suites is 0.0000, 0.0000, and 0.0072 for no-incoming-carry, incoming-carry, and thousands-carry cases, respectively. A matched extra-data control leaves that picture essentially unchanged at 0.0000, 0.0002, and 0.0116. By contrast, the targeted probe family rises to 0.0332, 0.0450, and 0.0250.

The gain is modest in absolute terms, but it is highly diagnostic. At the hundreds prediction step, the mean logit margin \(\Logit(\text{flag digit})-\Logit(\text{true digit})\) is strongly positive in both the base family and the matched control: \(+288.0/+98.9/+238.8\) for \(\family{base}\) and \(+335.9/+101.7/+271.5\) for \(\family{ctrl\_dup}\) on \(\family{probe\_h0}\), \(\family{probe\_h1}\), and \(\family{probe\_t1}\), respectively. Under targeted probes, the sign reverses to \(-109.9/-105.2/-30.1\). Attention redistributes at the same time: mean attention to lower-order digits falls from roughly 0.64--0.70 in the base/control families to 0.35--0.44 in the targeted family, while attention to upper positions rises. Taken together, these behavior, logit, and attention summaries are consistent with a carry-flag interpretation. The hundreds position initially behaves like a binary carry indicator, and the targeted repair works because of what it teaches, not because of extra optimization alone (Appendix Table~\ref{tab:carryprobe_appendix}).

This result also clarifies what comes next. The targeted probes repair the upper digits more than the lower tail. Under \family{probe\_all}, the probability of getting the top two digits correct rises sharply, but the probability of also getting both lower digits correct, given upper success, remains low. Carry repair therefore reveals the next barrier rather than removing it (Appendix Table~\ref{tab:upperlower_appendix}).

\subsection{After carry repair, the next bottleneck is conditional recomposition}

Once the carry-flag trap is weakened, the central question becomes what still limits exact match. The analytic binding ablation in Table~\ref{tab:binding_small} gives a clear answer. High-only data barely improves over the matched control. Tail-only data helps more. But the high-conditioned tail family, which teaches the lower tail in the context of the correct upper state, is best on all three true-3-digit suites.

The same ordering appears in the recomposition metric \(P(\mathrm{exact}\mid\mathrm{high2\ correct})\). Relative to the matched control, \family{tailhigh} raises this quantity by 0.291, 0.280, and 0.251 on the no-incoming-carry, incoming-carry, and thousands-carry suites, respectively. Relative to tail-only, it still improves by 0.192, 0.184, and 0.151. The clearest interpretation is that the late-stage bottleneck is not generic lower-digit noise. The model often already has the upper state it needs. What it still lacks is the ability to attach the correct lower tail to that upper state.

\begin{table}[t]
\centering
\small
\caption{\textbf{Analytic carry-repaired binding ablation (5 seeds; 1-layer, width-16).} All families are compared from the same carry-repaired base with a volume-matched final augmentation pack. The decisive ingredient is high-conditioned tail binding.}
\label{tab:binding_small}
\begin{tabular}{lccc}
\toprule
Family & No incoming carry & Incoming carry & Thousands-carry \\
\midrule
\multicolumn{4}{l}{\emph{Exact match}} \\
\family{ctrl3} & 0.310 & 0.416 & 0.390 \\
\family{highonly} & 0.326 & 0.414 & 0.379 \\
\family{tail} & 0.396 & 0.512 & 0.468 \\
\family{tailhigh} & \textbf{0.541} & \textbf{0.700} & \textbf{0.585} \\
\midrule
\multicolumn{4}{l}{\emph{\(P(\mathrm{exact}\mid\mathrm{high2\ correct})\)}} \\
\family{ctrl3} & 0.446 & 0.482 & 0.500 \\
\family{highonly} & 0.454 & 0.481 & 0.478 \\
\family{tail} & 0.545 & 0.578 & 0.600 \\
\family{tailhigh} & \textbf{0.736} & \textbf{0.763} & \textbf{0.751} \\
\bottomrule
\end{tabular}
\end{table}

The bridge experiment shows that this ordering is not confined to the analytic 1-layer setting. When the same conditional-binding comparison is repeated in the 2-layer, width-32 model used later for the residual study, the ranking \family{tailhigh} \(>\) \family{tail} \(>\) \(\max(\family{ctrl3},\family{highonly})\) reappears on all three suites. Exact-match means become 0.733, 0.753, and 0.716 for \family{tailhigh}, and \family{tailhigh} beats \family{tail} in all 5 paired seeds on every suite. The same is true for \(P(\mathrm{exact}\mid\mathrm{high2\ correct})\), where the large-capacity means reach 0.800, 0.801, and 0.787 (Appendix Table~\ref{tab:binding_large}). This bridge matters because it puts the recomposition claim on the same backbone capacity as the later tens-residual analysis.

The residual structure after \family{tailhigh} makes the story sharper still. In the large-capacity bridge, among examples whose top two digits are correct but whose full output is still wrong, 95--97\% are tens-only errors. By this point the failure is no longer a broad ``lower tail'' problem. It is a much narrower tens residual, which motivates the final stage.

\subsection{The final major residual is a carry-conditioned tens-sign error}

\begin{figure}[t]
\centering
\includegraphics[width=0.96\linewidth]{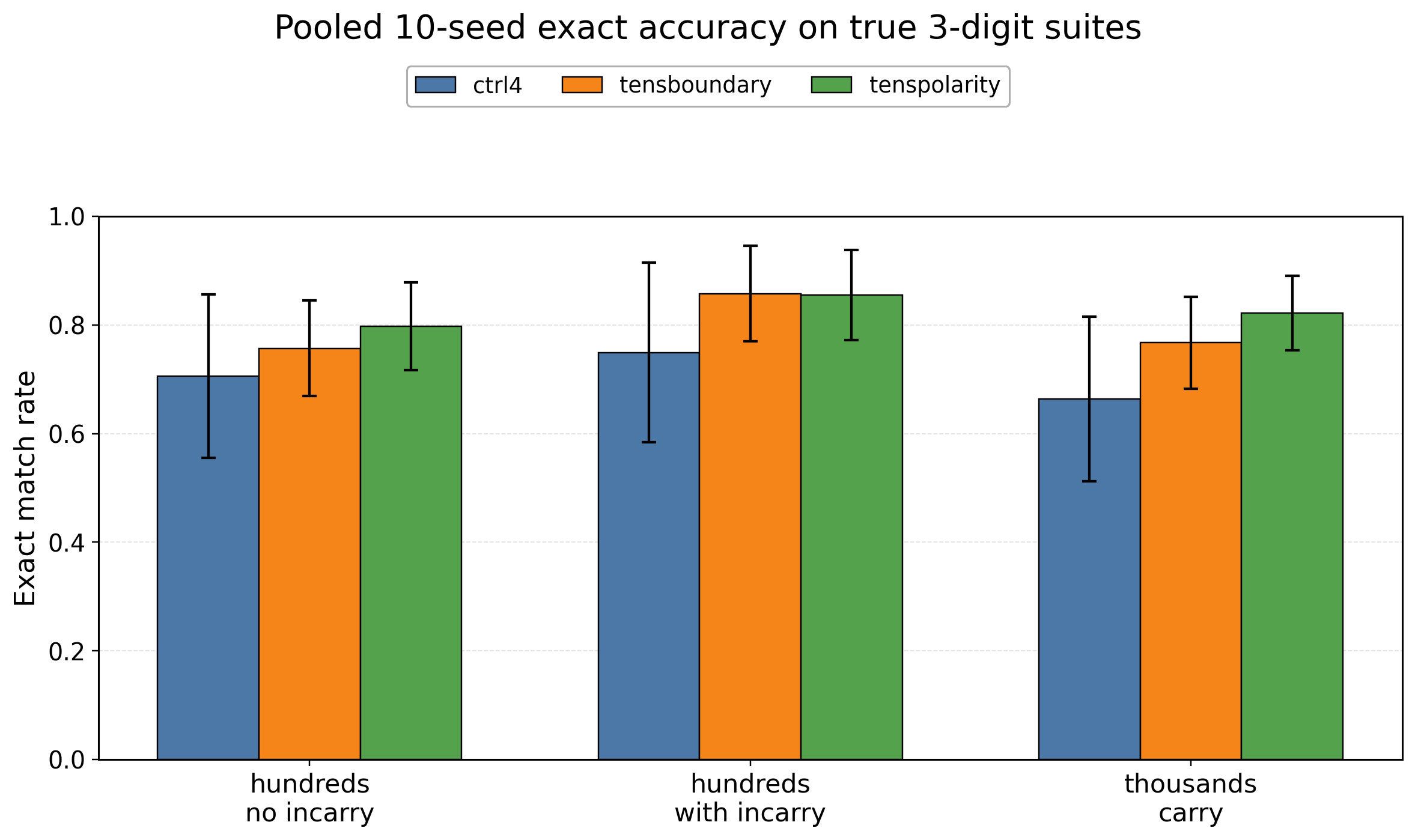}
\caption{\textbf{10-seed late-stage confirmation: pooled exact-match means on the three true-3-digit suites.} This is a downstream confirmation study in the larger 2-layer, width-32 model. \family{tenspolarity} is best on the no-incoming-carry and thousands-carry suites, while \family{tensboundary} is narrowly best on the incoming-carry suite.}
\label{fig:exact}
\end{figure}

\begin{figure}[t]
\centering
\includegraphics[width=0.96\linewidth]{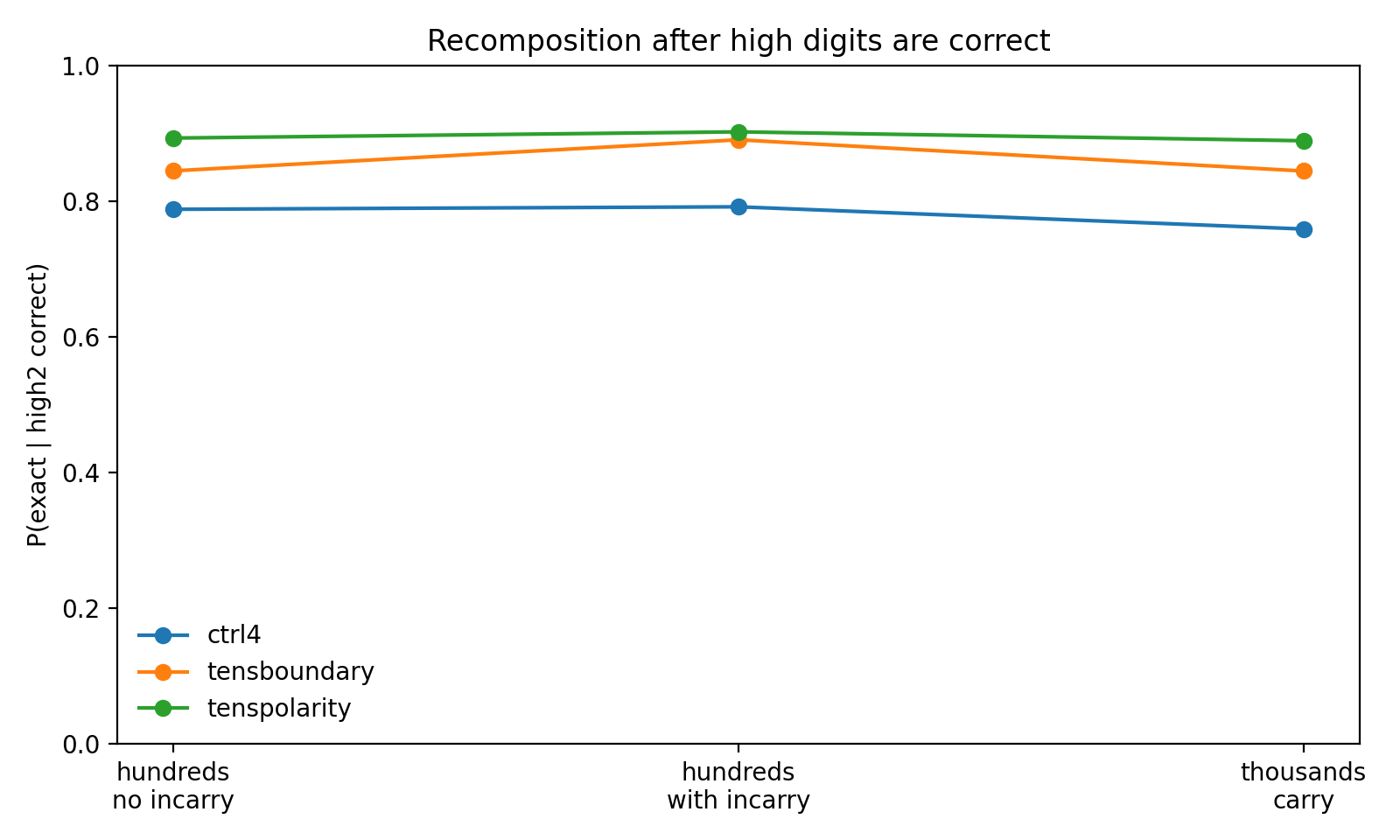}
\caption{\textbf{10-seed late-stage confirmation: recomposition after the upper digits are already correct.} The metric is \(P(\mathrm{exact}\mid\mathrm{high2\ correct})\). Improvements here are important because they show that later gains are not explained only by fixing the upper digits.}
\label{fig:recomposition10}
\end{figure}

The bridge experiment suggests that after recomposition succeeds, the remaining error should be highly localized. The separate 10-seed late-stage confirmation study tests that last residual directly. It compares three families: \family{ctrl4}, \family{tensboundary}, and \family{tenspolarity}. Figure~\ref{fig:exact} reports pooled exact-match means, Figure~\ref{fig:recomposition10} reports the corresponding recomposition metric, and Table~\ref{tab:pooled10} summarizes the exact means together with the strongest paired-seed evidence.

The strongest robustness claim is reserved for the hardest \suite{true3\_thousands\_carry} suite. There, \family{tenspolarity} improves exact match from 0.664 to 0.822 relative to \family{ctrl4}, wins on 9 of 10 paired seeds, and also beats \family{tensboundary} on 9 of 10 paired seeds. Figure~\ref{fig:recomposition10} shows that the later families also improve \(P(\mathrm{exact}\mid\mathrm{high2\ correct})\), which means that they continue to repair the answer even after the upper state is right.

\begin{figure}[t]
\centering
\includegraphics[width=0.96\linewidth]{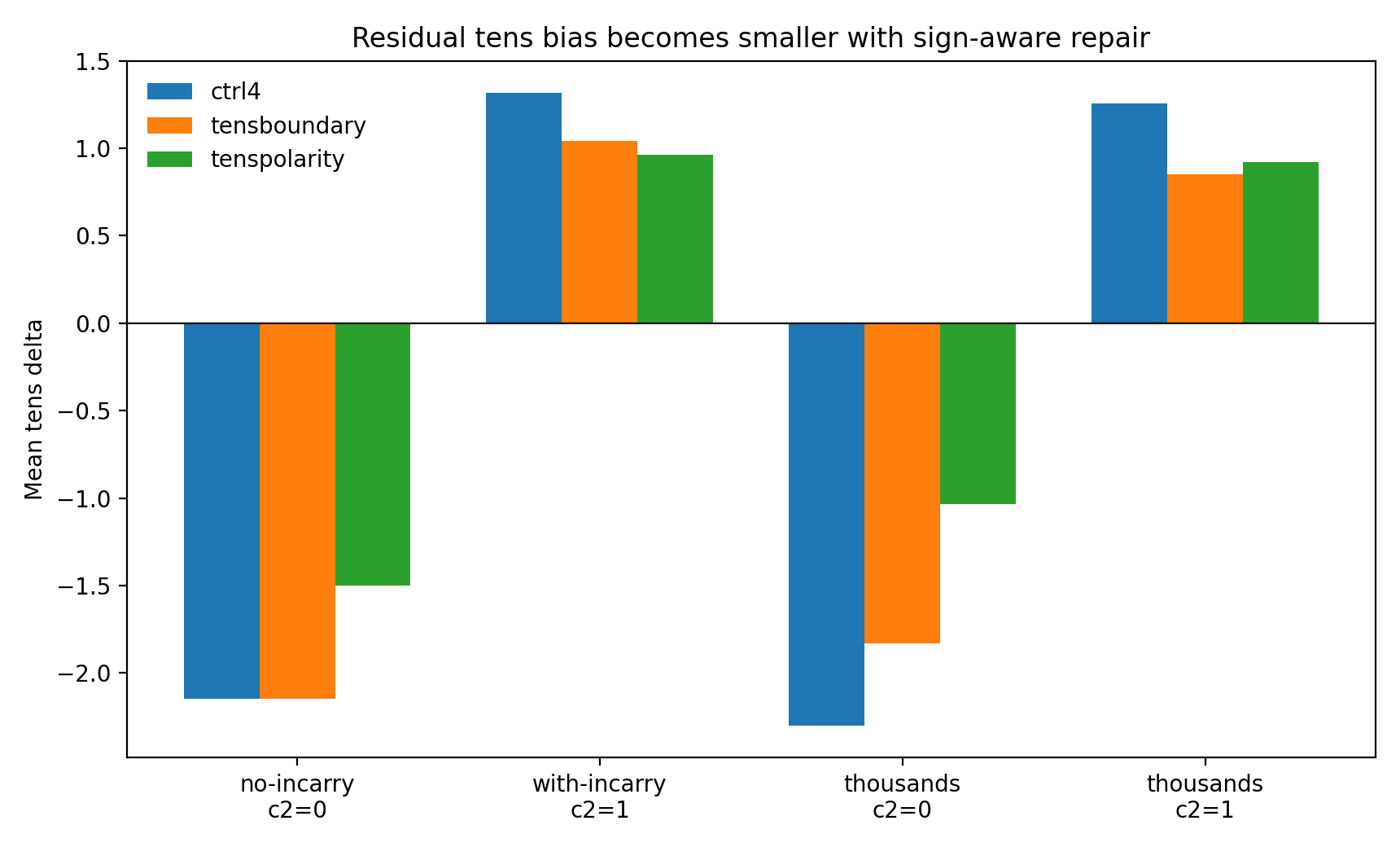}
\caption{\textbf{10-seed late-stage confirmation: signed tens residuals.} Negative values indicate systematic under-shooting of the tens digit; positive values indicate systematic over-shooting. \family{tenspolarity} moves both \(c_2=0\) and \(c_2=1\) branches closer to zero, most clearly on the hardest thousands-carry suite.}
\label{fig:tensbias}
\end{figure}

The signed tens-residual plot in Figure~\ref{fig:tensbias} explains why the improvement looks the way it does. When \(c_2=0\), the model tends to under-shoot the tens digit. When \(c_2=1\), it tends to over-shoot. \family{tenspolarity} contracts both branches toward zero, most clearly on the hardest thousands-carry suite. The final late-stage residual is therefore not generic lower-tail noise. It is a carry-conditioned tens-sign error that only becomes visible after the earlier barriers have already been weakened.

\begin{table}[t]
\centering
\small
\caption{\textbf{10-seed late-stage confirmation means and strongest paired evidence.} The strongest paired claim is reserved for the hardest thousands-carry suite.}
\label{tab:pooled10}
\begin{tabular}{lccc}
\toprule
Family & No incoming carry & Incoming carry & Thousands-carry \\
\midrule
\family{ctrl4} & 0.7057 & 0.7492 & 0.6640 \\
\family{tensboundary} & 0.7572 & \textbf{0.8579} & 0.7674 \\
\family{tenspolarity} & \textbf{0.7973} & 0.8550 & \textbf{0.8219} \\
\bottomrule
\end{tabular}
\vspace{0.5em}

\begin{tabular}{lcc}
\toprule
Paired comparison on \suite{true3\_thousands\_carry} & Wins & Mean delta \\
\midrule
\family{tenspolarity} - \family{ctrl4} & 9/10 & +0.1579 \\
\family{tenspolarity} - \family{tensboundary} & 9/10 & +0.0545 \\
\bottomrule
\end{tabular}
\end{table}

\section{Discussion}

The central result is that arithmetic OOD failure in this minimal GPT is sequential and experimentally separable. Even though the 2-digit training set already covers all local digit transitions, the model still fails in a consistent order: it first loses the mapping between layout and role, then assigns the wrong meaning to the hundreds position, and then struggles to bind the correct lower tail to an otherwise correct upper state. Only after those broader failures shrink does the final tens residual become visible.

This decomposition changes how later repairs should be interpreted. Mixed-layout exposure matters because it weakens a layout barrier. The targeted carry probes matter because they alter a specific wrong semantic rule that a matched extra-data control leaves intact. High-conditioned tail data matters because it addresses the binding problem that remains after carry repair. The late tens-residual families matter because they act on a narrow, carry-conditioned branch error rather than on a generic lower-digit confusion. In each case, the intervention is informative because it is matched to a concrete hypothesis about the active barrier.

The broader methodological lesson is that arithmetic benchmarks are most useful when each new suite adds one new source of difficulty. Otherwise, format robustness, semantic correctness, and recomposition are compressed into a single score. The same logic should extend beyond arithmetic wherever local rule coverage is easy to saturate but correct recomposition across positions remains difficult. This preference for controlled, inspectable, and auditable workflows also aligns with recent calls for black-box-free AI practice and deterministic quality control in self-hosted local LLM systems \citep{shintani2026lecture2quiz,shintani2025ai2l}.

\section{Limitations}

The study still concerns a deliberately small and synthetic system. The position-family variants are simplified inspired-by comparisons rather than exact reproductions of previously published systems. The carry-flag claim is supported by behavior, logits, and attention summaries, but not by a full weight-level circuit proof. The new 2-layer bridge reduces the capacity gap between the analytic binding study and the late-stage residual study, but we do not yet provide a full grid over model size, optimizer settings, or alternative arithmetic formats.

\section{Conclusion}

The value of an arithmetic benchmark is not only in whether a model passes it, but also in whether the benchmark lets us say what failed and why. We show that arithmetic OOD failure in this controlled minimal GPT is not a single phenomenon. It unfolds in four experimentally separable stages: a layout barrier, a carry-flag semantic error, a conditional recomposition bottleneck, and a final carry-conditioned tens residual. Because each stage can be weakened by a matched intervention, this decomposition is not merely descriptive; it is experimentally testable. We therefore argue for using arithmetic benchmarks not only as single-number contests, but as controlled settings in which distinct computational barriers can be exposed and studied.

\section*{Data and code availability}

All experiments in this study use fully synthetic data generated on the fly from explicit arithmetic rules; no external dataset is required. The arXiv submission includes reader-facing ancillary files under \path{anc/}. The index file \path{anc/README_companion_materials.md} lists five standalone Colab notebooks, one for each main experiment family in the paper, together with five compact result archives carrying the same experiment-level names. The compact archives retain only the summary files needed to check the manuscript; exploratory notebooks, duplicate variants, model checkpoints, and large intermediate logs are intentionally omitted. The standalone notebooks regenerate the synthetic training and evaluation data and can reproduce the reader-facing result packages directly in Google Colab.

\section*{Acknowledgements}

This work was supported by JSPS KAKENHI Grant Number JP25K00269 (Grant-in-Aid for Scientific Research (C), Project Title: ``Elucidation of Myosin Molecular Dynamics Associated with Sarcomere Morphological Changes in the Intracellular Environment''), by Chubu University FY2025 Special Research Fund (CP) (``Development of an AI safety evaluation and AI2L-based utilization system for clinical laboratory technologist education''), and by the Chubu University FY2025 Research Institute for Industry and Economics (RIIE) Research Project (``Development and evaluation of an explainable management-analysis support tool using local small-scale LLMs'').

The author used ChatGPT (OpenAI) for language polishing during manuscript preparation, reviewed the resulting text, and takes full responsibility for the manuscript.

\appendix

\section{Matched carry-probe diagnostics}

Table~\ref{tab:carryprobe_appendix} collects the matched carry-probe summary used in Section~4.2. Positive margins mean that the model favors the flag digit over the true digit at the relevant prediction step.

\begin{table}[ht]
\centering
\small
\caption{\textbf{Matched carry-probe summary (5 seeds).} Exact-match means are shown for the three true-3-digit suites. Positive margins indicate preference for the flag digit over the true digit at the corresponding probe step.}
\label{tab:carryprobe_appendix}
\begin{tabular}{lcccccc}
\toprule
Family & No-inc. exact & Inc. exact & 1000-carry exact & \(\Delta_{h0}\) & \(\Delta_{h1}\) & \(\Delta_{t1}\) \\
\midrule
\family{base} & 0.0000 & 0.0000 & 0.0072 & +288.0 & +98.9 & +238.8 \\
\family{ctrl\_dup} & 0.0000 & 0.0002 & 0.0116 & +335.9 & +101.7 & +271.5 \\
\family{probe\_all} & 0.0332 & 0.0450 & 0.0250 & -109.9 & -105.2 & -30.1 \\
\bottomrule
\end{tabular}
\end{table}

\section{Carry repair improves upper digits before lower recomposition}

Table~\ref{tab:upperlower_appendix} shows why carry repair alone is not the end of the story. Under \family{probe\_all}, the probability of getting the upper digits correct rises sharply, but the probability of also getting both lower digits correct given upper success remains low. This is the pattern expected if the next barrier is recomposition rather than carry semantics.

\begin{table}[ht]
\centering
\small
\caption{\textbf{Upper-digit repair versus lower-digit recomposition (5 seeds).}}
\label{tab:upperlower_appendix}
\begin{tabular}{llcc}
\toprule
Family & Suite & Upper-correct rate & Lower-both-correct given upper \\
\midrule
\family{base} & No incoming carry & 0.0002 & 0.0000 \\
\family{base} & Incoming carry & 0.0000 & --- \\
\family{base} & Thousands-carry & 0.0406 & 0.1773 \\
\family{ctrl\_dup} & No incoming carry & 0.0012 & 0.0000 \\
\family{ctrl\_dup} & Incoming carry & 0.0014 & 0.1429 \\
\family{ctrl\_dup} & Thousands-carry & 0.0208 & 0.4231 \\
\family{probe\_all} & No incoming carry & 0.6888 & 0.0482 \\
\family{probe\_all} & Incoming carry & 0.5636 & 0.0788 \\
\family{probe\_all} & Thousands-carry & 0.3050 & 0.0813 \\
\bottomrule
\end{tabular}
\end{table}

\section{Large-capacity bridge of the conditional-binding order}

Table~\ref{tab:binding_large} repeats the conditional-binding comparison at the same capacity as the late-stage tens-residual study (2-layer, width-32). The ordering is the same as in the analytic 1-layer study: \family{tailhigh} is best on both exact match and \(P(\mathrm{exact}\mid\mathrm{high2\ correct})\), and it beats \family{tail} in all 5 paired seeds on every suite for both metrics.

\begin{table}[ht]
\centering
\small
\caption{\textbf{Large-capacity bridge of the conditional-binding ablation (5 seeds; 2-layer, width-32).}}
\label{tab:binding_large}
\begin{tabular}{lccc}
\toprule
Family & No incoming carry & Incoming carry & Thousands-carry \\
\midrule
\multicolumn{4}{l}{\emph{Exact match}} \\
\family{ctrl3} & 0.418 & 0.496 & 0.458 \\
\family{highonly} & 0.456 & 0.540 & 0.476 \\
\family{tail} & 0.535 & 0.568 & 0.530 \\
\family{tailhigh} & \textbf{0.733} & \textbf{0.753} & \textbf{0.716} \\
\midrule
\multicolumn{4}{l}{\emph{\(P(\mathrm{exact}\mid\mathrm{high2\ correct})\)}} \\
\family{ctrl3} & 0.524 & 0.542 & 0.546 \\
\family{highonly} & 0.577 & 0.587 & 0.577 \\
\family{tail} & 0.642 & 0.617 & 0.617 \\
\family{tailhigh} & \textbf{0.800} & \textbf{0.801} & \textbf{0.787} \\
\bottomrule
\end{tabular}
\end{table}

\section{Additional pooled metrics from the late-stage 10-seed study}

Table~\ref{tab:recomp_appendix} reports the pooled recomposition summary for the 10-seed late-stage confirmation. These numbers make explicit that the late-stage gains are not confined to exact match alone.

\begin{table}[ht]
\centering
\small
\caption{\textbf{Pooled 10-seed recomposition summary.}}
\label{tab:recomp_appendix}
\begin{tabular}{llccc}
\toprule
Family & Suite & \texttt{high2\_correct} & \texttt{low2\_correct} & $P(\mathrm{exact}\mid\mathrm{high2\ correct})$ \\
\midrule
\family{ctrl4} & No incoming carry & 0.8942 & 0.7650 & 0.7878 \\
\family{tensboundary} & No incoming carry & 0.8968 & 0.8289 & 0.8445 \\
\family{tenspolarity} & No incoming carry & 0.8931 & 0.8819 & 0.8928 \\
\family{ctrl4} & Incoming carry & 0.9424 & 0.7843 & 0.7915 \\
\family{tensboundary} & Incoming carry & 0.9625 & 0.8889 & 0.8903 \\
\family{tenspolarity} & Incoming carry & 0.9474 & 0.9007 & 0.9020 \\
\family{ctrl4} & Thousands-carry & 0.8731 & 0.7378 & 0.7588 \\
\family{tensboundary} & Thousands-carry & 0.9096 & 0.8341 & 0.8442 \\
\family{tenspolarity} & Thousands-carry & 0.9252 & 0.8822 & 0.8889 \\
\bottomrule
\end{tabular}
\end{table}

\section{Additional signed tens-residual summary}

Table~\ref{tab:tensbias_appendix} reports the mean signed tens deltas behind Figure~\ref{fig:tensbias}. Negative values denote under-shooting; positive values denote over-shooting.

\begin{table}[ht]
\centering
\small
\caption{\textbf{Mean signed tens residuals in the pooled 10-seed confirmation.}}
\label{tab:tensbias_appendix}
\begin{tabular}{llcc}
\toprule
Family & Suite branch & Mean tens delta & Interpretation \\
\midrule
\family{ctrl4} & No incoming carry ($c_2=0$) & -2.149 & under-shoot \\
\family{tensboundary} & No incoming carry ($c_2=0$) & -2.148 & under-shoot \\
\family{tenspolarity} & No incoming carry ($c_2=0$) & -1.499 & reduced under-shoot \\
\family{ctrl4} & Incoming carry ($c_2=1$) & +1.319 & over-shoot \\
\family{tensboundary} & Incoming carry ($c_2=1$) & +1.044 & reduced over-shoot \\
\family{tenspolarity} & Incoming carry ($c_2=1$) & +0.961 & further reduced over-shoot \\
\family{ctrl4} & Thousands-carry, $c_2=0$ & -2.302 & under-shoot \\
\family{tensboundary} & Thousands-carry, $c_2=0$ & -1.829 & reduced under-shoot \\
\family{tenspolarity} & Thousands-carry, $c_2=0$ & -1.034 & strongly reduced under-shoot \\
\family{ctrl4} & Thousands-carry, $c_2=1$ & +1.258 & over-shoot \\
\family{tensboundary} & Thousands-carry, $c_2=1$ & +0.854 & reduced over-shoot \\
\family{tenspolarity} & Thousands-carry, $c_2=1$ & +0.923 & reduced over-shoot \\
\bottomrule
\end{tabular}
\end{table}

\bibliographystyle{plainnat}
\bibliography{references}

@misc{karpathy2026microgpt,
  author = {Andrej Karpathy},
  title = {{microgpt}},
  year = {2026},
  month = feb,
  url = {https://karpathy.github.io/2026/02/12/microgpt/},
  note = {Blog post}
}

@inproceedings{vaswani2017attention,
  author = {Ashish Vaswani and Noam Shazeer and Niki Parmar and Jakob Uszkoreit and Llion Jones and Aidan N. Gomez and {\L}ukasz Kaiser and Illia Polosukhin},
  title = {Attention Is All You Need},
  booktitle = {Advances in Neural Information Processing Systems},
  volume = {30},
  year = {2017}
}

@inproceedings{shaw2018relative,
  author = {Peter Shaw and Jakob Uszkoreit and Ashish Vaswani},
  title = {Self-Attention with Relative Position Representations},
  booktitle = {Proceedings of the 2018 Conference of the North American Chapter of the Association for Computational Linguistics: Human Language Technologies, Volume 2 (Short Papers)},
  pages = {464--468},
  publisher = {Association for Computational Linguistics},
  year = {2018}
}

@article{power2022grokking,
  author = {Alethea Power and Yuri Burda and Harri Edwards and Igor Babuschkin and Vedant Misra},
  title = {Grokking: Generalization Beyond Overfitting on Small Algorithmic Datasets},
  journal = {arXiv preprint arXiv:2201.02177},
  year = {2022}
}

@inproceedings{nanda2023grokkingmi,
  author = {Neel Nanda and Lawrence Chan and Tom Lieberum and Jess Smith and Jacob Steinhardt},
  title = {Progress Measures for Grokking via Mechanistic Interpretability},
  booktitle = {The Eleventh International Conference on Learning Representations},
  year = {2023}
}

@inproceedings{quirke2024addition,
  author = {Philip Quirke and Fazl Barez},
  title = {Understanding Addition in Transformers},
  booktitle = {The Twelfth International Conference on Learning Representations},
  year = {2024}
}

@article{quirke2024arithmetic,
  author = {Philip Quirke and Clement Neo and Fazl Barez},
  title = {Arithmetic in Transformers Explained},
  journal = {arXiv preprint arXiv:2402.02619},
  year = {2024}
}

@inproceedings{lee2024teaching,
  author = {Nayoung Lee and Kartik Sreenivasan and Jason D. Lee and Kangwook Lee and Dimitris Papailiopoulos},
  title = {Teaching Arithmetic to Small Transformers},
  booktitle = {The Twelfth International Conference on Learning Representations},
  year = {2024}
}

@article{zhou2024lengthnotrobust,
  author = {Yongchao Zhou and Uri Alon and Xinyun Chen and Xuezhi Wang and Rishabh Agarwal and Denny Zhou},
  title = {Transformers Can Achieve Length Generalization But Not Robustly},
  journal = {arXiv preprint arXiv:2402.09371},
  year = {2024}
}

@inproceedings{cho2024position,
  author = {Hanseul Cho and Jaeyoung Cha and Pranjal Awasthi and Srinadh Bhojanapalli and Anupam Gupta and Chulhee Yun},
  title = {Position Coupling: Improving Length Generalization of Arithmetic Transformers Using Task Structure},
  booktitle = {Advances in Neural Information Processing Systems},
  volume = {37},
  year = {2024}
}

@inproceedings{mcleish2024abacus,
  author = {Sean McLeish and Arpit Bansal and Alex Stein and Neel Jain and John Kirchenbauer and Brian R. Bartoldson and Bhavya Kailkhura and Abhinav Bhatele and Jonas Geiping and Avi Schwarzschild and Tom Goldstein},
  title = {Transformers Can Do Arithmetic with the Right Embeddings},
  booktitle = {Advances in Neural Information Processing Systems},
  volume = {37},
  year = {2024}
}

@article{sabbaghi2024symmetry,
  author = {Mahdi Sabbaghi and George Pappas and Hamed Hassani and Surbhi Goel},
  title = {Explicitly Encoding Structural Symmetry is Key to Length Generalization in Arithmetic Tasks},
  journal = {arXiv preprint arXiv:2406.01895},
  year = {2024}
}

@inproceedings{huang2025formal,
  author = {Xinting Huang and Andy Yang and Satwik Bhattamishra and Yash Sarrof and Andreas Krebs and Hattie Zhou and Preetum Nakkiran and Michael Hahn},
  title = {A Formal Framework for Understanding Length Generalization in Transformers},
  booktitle = {The Thirteenth International Conference on Learning Representations},
  year = {2025}
}

@inproceedings{cho2025operandcount,
  author = {Hanseul Cho and Jaeyoung Cha and Srinadh Bhojanapalli and Chulhee Yun},
  title = {Arithmetic Transformers Can Length-Generalize in Both Operand Length and Count},
  booktitle = {The Thirteenth International Conference on Learning Representations},
  year = {2025}
}

@inproceedings{yan2025additiondiagnostic,
  author = {Yang Yan and Yu Lu and Renjun Xu and Zhenzhong Lan},
  title = {Do Large Language Models Truly Grasp Addition? A Rule-Focused Diagnostic Using Two-Integer Arithmetic},
  booktitle = {Proceedings of the 2025 Conference on Empirical Methods in Natural Language Processing},
  pages = {13467--13483},
  publisher = {Association for Computational Linguistics},
  year = {2025},
  doi = {10.18653/v1/2025.emnlp-main.681}
}

@inproceedings{abbe2024scratchpad,
  author = {Emmanuel Abbe and Samy Bengio and Aryo Lotfi and Colin Sandon and Omid Saremi},
  title = {How Far Can Transformers Reason? The Globality Barrier and Inductive Scratchpad},
  booktitle = {Advances in Neural Information Processing Systems},
  volume = {37},
  year = {2024}
}

@article{zaremba2014execute,
  author = {Wojciech Zaremba and Ilya Sutskever},
  title = {Learning to Execute},
  journal = {arXiv preprint arXiv:1410.4615},
  year = {2014}
}

@inproceedings{kaiser2015neuralgpu,
  author = {{\L}ukasz Kaiser and Ilya Sutskever},
  title = {Neural {GPUs} Learn Algorithms},
  booktitle = {The Fourth International Conference on Learning Representations},
  year = {2016},
  url = {https://openreview.net/forum?id=ryjp1c9xg}
}

@inproceedings{dehghani2019universal,
  author = {Mostafa Dehghani and Stephan Gouws and Oriol Vinyals and Jakob Uszkoreit and {\L}ukasz Kaiser},
  title = {Universal Transformers},
  booktitle = {The Seventh International Conference on Learning Representations},
  year = {2019},
  url = {https://openreview.net/forum?id=HyzdRiR9Y7}
}

@inproceedings{press2022alibi,
  author = {Ofir Press and Noah A. Smith and Mike Lewis},
  title = {Train Short, Test Long: Attention with Linear Biases Enables Input Length Extrapolation},
  booktitle = {The Tenth International Conference on Learning Representations},
  year = {2022},
  url = {https://openreview.net/forum?id=R8sQPpGCv0}
}

@article{shintani2026lecture2quiz,
  author = {Seine A. Shintani},
  title = {Self-hosted Lecture-to-Quiz: Local {LLM} {MCQ} Generation with Deterministic Quality Control},
  journal = {arXiv preprint arXiv:2603.08729},
  year = {2026},
  doi = {10.48550/arXiv.2603.08729},
  url = {https://arxiv.org/abs/2603.08729}
}

@article{shintani2025ai2l,
  author = {Seine A. Shintani},
  title = {{AI} to Learn ({AI2L}): Human-Centered Guidelines for {Black-Box-Free} {AI} and Empirical Law Discovery via Symbolic Regression},
  journal = {Jxiv preprint},
  year = {2025},
  doi = {10.51094/jxiv.1435},
  url = {https://doi.org/10.51094/jxiv.1435},
  note = {Version 2}
}

\end{document}